\documentclass[11pt,a4paper]{article}
\usepackage[hyperref]{acl2021}
\usepackage{times}
\usepackage{latexsym}

\usepackage{multirow}
\usepackage{graphicx}
\usepackage{tabularx}

\usepackage{array}
\newcolumntype{M}[1]{>{\centering\arraybackslash}m{#1}}
\newcolumntype{C}[1]{>{\centering\arraybackslash}m{#1}}

\usepackage{microtype}

\aclfinalcopy 
\def\aclpaperid{***} 

\title{Biomedical Data-to-Text Generation via Fine-Tuning Transformers}

\date{}

\author{Ruslan Yermakov \\
  Decision Science  \\
 \& Advanced Analytics \\
  Bayer AG\\
  \texttt{yermakovruslan@gmail.com} 
\\\And
  Nicholas Drago \\
  Regulatory Policy \\
  and Intelligence  \\
  Bayer AG\\
\\\And
  Angelo Ziletti* \\
  Decision Science  \\
 \& Advanced Analytics \\
  Bayer AG\\
  \texttt{angelo.ziletti@bayer.com} \\}

\begin{document}
\maketitle
\begin{abstract}

Data-to-text (D2T) generation in the biomedical domain is a promising - yet mostly unexplored - field of research.
Here, we apply neural models for D2T generation to a real-world dataset consisting of package leaflets of European medicines.
We show that fine-tuned transformers are able to generate realistic, multi-sentence text from data in the biomedical domain, yet have important limitations.
We also release a new dataset (\emph{BioLeaflets}) for benchmarking D2T generation models in the biomedical domain. 
\end{abstract}

\section{Introduction} \label{sec:Intro}

Data-to-text (D2T) systems are attracting considerable interest due to their ability to automate the time-consuming writing of data-driven reports. 
There is a hitherto largely untapped potential for text generation in the biomedical domain. Potential applications of natural language generation of patient-friendly biomedical text include preparation of the first draft of package leaflets, patient education materials, or direct-to-consumer promotional materials in countries where this is permitted.
Here we focus on a D2T task aiming to generate fluent and fact-based descriptions from biomedical data. 

\section{Related Work} \label{sec:Related}

Recently, neural D2T models have significantly improved the quality of short text generation (usually one sentence long) from input data compared to multi-stage pipelined or template-based approaches. 
Examples include biographies from Wikipedia fact tables ~\citep{Lebret:16}, restaurant descriptions from meaning representations ~\citep{Novikova-e2e:17}, and basketball game summaries from statistical tables ~\citep{Wiseman:17}. 
Still, neural D2T approaches have major challenges, as outlined by ~\citet{Wiseman:17} and ~\citet{Parikh:20} which hinder their application to many real-world applications. 
These include hallucination effects (generated phrases not supported or contradictory to the source data), missing facts (generated text does not include input information), intersentence incoherence, and repetitiveness in the generated text. 
Following the success of leveraging pre-trained large-scale language models for a large variety of tasks, 
\citet{Kale:20} fine-tuned T5 models ~\citep{T5_model:20} for D2T generation. 
This strategy achieved state-of-the-art performance on task-oriented dialogue (MultiWoz) ~\citep{Multiwoz:18}, tables-to-text (ToTTo) ~\citep{Parikh:20} and graph-to-text (WebNLG) ~\citep{Webnlg:17}. 

To the best of our knowledge, recent neural approaches and transfer learning strategies have not been applied to multi-sentence generation from input data, nor have they been applied in the biomedical domain. 
Our contribution is two-fold: we introduce a real-world biomedical dataset \emph{BioLeaflets}, and demonstrate that transformers can generate high-quality multi-sentence text from data in the biomedical domain. 
The \emph{BioLeaflets} dataset, fine-tuned models, code, and generated samples are available at \url{https://github.com/bayer-science-for-a-better-life/data2text-bioleaflets}.

\begin{table*}[ht!]
\small
\centering
\begin{tabular}{ | p{1.5cm} | p{14cm}| } 

 \hline
 
 (a) \newline Original section content & \textbf{novonorm} is \textbf{an oral antidiabetic medicine} containing \textbf{repaglinide} which helps your \textbf{pancreas} produce more \textbf{insulin} and thereby lower \textbf{your blood sugar} (\textbf{glucose}). \textbf{type 2 diabetes} is \textbf{a disease} in which your \textbf{pancreas} does not make enough \textbf{insulin} to control \textbf{the sugar} in your blood or where your \textbf{body} does not respond normally to the \textbf{insulin} it produces. \textbf{novonorm} is used to control \textbf{type 2 diabetes} in adults as an add-on to diet and exercise: \textbf{treatment} is usually started if diet, \textbf{exercise and weight reduction} alone have not been able to control (or lower) \textbf{your blood sugar}. \textbf{novonorm} can also be given with \textbf{metformin}, another medicine for \textbf{diabetes}. \textbf{novonorm} has been shown to lower \textbf{the blood sugar}, which helps to prevent \textbf{complications} from \textbf{your diabetes}. \\

 \hline
 
  (b) \newline Input: \newline entities as a flat string & \verb|<PRODUCT_NAME> novonorm </PRODUCT_NAME> <TREATMENT>| \newline
 \verb|an_oral_antidiabetic_medicine </TREATMENT> <GENERIC_NAME> repaglinide |
 \newline 
 \verb|</GENERIC_NAME> <SYSTEM_ORGAN_SITE> pancreas </SYSTEM_ORGAN_SITE>|
 \newline  
 \verb| <GENERIC_NAME> insulin </GENERIC_NAME> <TEST> your_blood_sugar </TEST>| 
 \newline 
 \verb| | \ldots \verb|<TEST> the_blood_sugar </TEST> <DX_NAME> complications </DX_NAME>| 
 \newline 
 \verb| <PROBLEM> your_diabetes </PROBLEM>| \\

 \hline
 
 (c) \newline T5 generated section & \textbf{novonorm} is \textbf{an oral antidiabetic medicine} containing \textbf{repaglinide} which helps your \textbf{pancreas} produce more \textbf{insulin} and thereby lower \textbf{your blood sugar} (\textbf{glucose}). \textbf{type 2 diabetes} is \textbf{a disease} in which your \underline{panccreases} does not make enough \textbf{insulin} to control \textbf{the sugar} in your blood or where your \textbf{body} does not respond normally to the \textbf{insulin} it produces. \underline{novonourm} is used to control \textcolor{red}{type 2 diabetic patients who still have their \textbf{diabetes} and who are still} unable to respond normally despite \textbf{treatment}, \textbf{exercise and weight reduction}, which have not helped to control your blood \textbf{glucose}. \textcolor{red}{there is also a test that} \underline{novonorgm} can be given in combination with \textbf{metformin}, another medicine for \textbf{diabetes}. it is important that you continue to follow the advice on diet, exercising and diet \textcolor{red}{that your nurse or doctor has given you}. \\
 
 \hline
 
 (d) \newline BART generated section & \underline{novondorm} is \textbf{an oral antidiabetic medicine} containing \textbf{repaglinide} which helps your \textbf{pancreas} produce more \textbf{insulin} and thereby lower \textbf{your blood sugar} (\textbf{glucose}). \textbf{type 2 diabetes} is \textbf{a disease} in which your \underline{pancrea} does not make enough \textbf{insulin} to control \textbf{the sugar} in your blood or where your \textbf{body} does not respond normally to the \textbf{insulin} it produces. \textbf{novonorm} is used to treat type 2 \underline{diabetic} in adults. \textcolor{red}{\textbf{treatment} is usually started with diet and \textbf{exercise and weight reduction}}. \textcolor{red}{your blood \textbf{glucose} may be increased when you start to take \underline{novonor} on its own or in combination with \textbf{metformin}. if you have \textbf{diabetes}, no} \\

 \hline
 
 (e) \newline Content planner generated section & \textbf{novonorm} contains the active substance \textbf{repaglinide} which helps to lower \textbf{your blood sugar} (\textbf{glucose}). \textbf{type 2 diabetes} is \textbf{a disease} where your \textbf{body} does not make enough \textbf{insulin} to control \textbf{the sugar} in your blood or where your \textbf{body} does not respond normally to the \textbf{insulin} it produces. \textcolor{red}{repaglinide krka is used to control \textbf{type 2 diabetes} in adults as \textbf{type 2 diabetes}.} \textcolor{red}{\textbf{type 2 diabetes} is also called non - insulin - dependent diabetes mellitus.} \textbf{type 2 diabetes} is also a condition in which your \textbf{body} does not make enough \textbf{insulin} \textcolor{red}{or the \textbf{insulin} that your \textbf{body} produces does not work as well as it should.} \textcolor{red}{your \textbf{body} can also make too much sugar. when this happens, sugar (\textbf{glucose}) builds up in the blood. this can lead to serious medical problems like heart disease, kidney disease, 2 and 2.} \\ 
 
 \hline

\end{tabular}
\caption{\label{generations-table} Example of text generations. Entities are highlighted in bold, typos are underlined, and hallucinations are shown in red.}
\end{table*}

\section{The \emph{BioLeaflets} Dataset} \label{sec:Dataset}

We introduce a new biomedical dataset for D2T generation - \emph{BioLeaflets}, a corpus of 1336 package leaflets of medicines authorised in Europe, which we obtain by scraping the European Medicines Agency (EMA) website. This dataset comprises the large majority ($\sim$ 90\%) of medicinal products authorised through the centralised procedure in Europe as of January 2021. 

Package leaflets are published for medicinal products approved in the European Union (EU). They are included in the packaging of medicinal products and contain information to help patients use the product safely and appropriately, under the guidance of their healthcare professional. Package leaflets are required to be written in a way that is clear and understandable \cite{PIL-2001}.
Each document contains six sections (see Table \ref{dataset-table}).The main challenges of this dataset for D2T generation are multi-sentence and multi-section target text,  small sample size, specialized medical vocabulary and syntax.

\subsection{Dataset Construction}
The content of each section is not standardized, yet it is still well-structured. 
Thus, we identify sections via heuristics such as regular expressions and word overlap. 
The content of each section is lower-cased and tokenized by treating all special characters as separate tokens. Duplicates are also removed.
We randomly split the dataset into training (80\%), development (10\%), and test (10\%) set. 
Table \ref{dataset-table} summarizes dataset statistics.

\begin{table*}[ht!]
\centering
\begin{tabular}{C{0.3\textwidth} C{0.08\textwidth} C{0.12\textwidth} C{0.12\textwidth} C{0.12\textwidth} C{0.12\textwidth}}
\hline
\hline
Section type & No. samples & Average length (characters) & Average length (tokens) & Average no. entities per section & No. unique entities \\

\hline
1. What the product is and what it is used for & $1\,314$ & $963$ & $174$ & $29.3$ & $9\,641$ \\
2. What you need to know before you take the product & $1\,309$  & $4\,560$ & $849$ & $127.7$ & $23\,278$  \\
3. How to take the product  & $1\,313$ & $2\,300$ & $458$ & $50.5$ & $11\,640$ \\
4. Possible side effects & $1\,295$ & $3\,453$ & $651$ & $135.2$ & $27\,945$  \\
5. How to store the product & $1\,172$ & $631$ & $123$ &  $6.3$ & $2\,041$ \\
6. Content of the pack and other information & $1\,311$ & $982$ & $196$ & $38.4$ & $9\,932$ \\
\hline
\hline
\end{tabular}
\caption{\label{dataset-table} \emph{BioLeaflets} dataset statistics grouped by section type.}
\end{table*}

\subsection{Dataset Annotations}

We do not have annotations available for the package leaflet text. To create the required input for D2T generation, we augment each document by leveraging named entity recognition (NER). 
\citet{Parikh:20} indicated it is important that target summaries contain information that can be inferred from the input data to avoid dataset-induced hallucinations. 
To this end, we combine two NER frameworks: Amazon Comprehend Medical (ACM) ~\citep{AWS_Comprehend:19} and Stanford Stanza ~\citep{Stanza:20,zhang2021biomedical}. 
ACM and Stanza achieved entity micro-averaged test F1 of 85.5\% and 88.13\% respectively on the 2010 i2b2/VA clinical dataset ~\citep{i2b2_dataset}. 
We further leverage ACM to detect medical conditions from ICD-10 \citep{icd-10} and medications from RxNorm.
Additionally, we treat all digits as entities, and add the medicine name as first entity. 
In case of overlapping entities from different sources, we favor longer entities over shorter ones. 
As a result of the NER process, we obtain 26 unique entity types. Examples are: \emph{problem}: ('active chronic hepatitis', 
'migraine pain'), \emph{system-organ-site}: ('blood vessel', 'kidneys', 'surrounding tissue'),  \emph{treatment}: ('routine dental care', 'a vaccination',  'a chemotherapy medicinal product'), or \emph{procedure}: ('injections', 'spinal or epidural anaesthesia', 'surgical intervention', 'bone marrow or stem cell transplant'). 

\emph{BioLeaflets} proposes a conditional generation task: given an ordered set of entities as source, the goal is to produce a multi-sentence section.
Since only the entities are provided as input, the structured data is underspecified. 
A human without specialized knowledge would likely be unable to produce satisfactory text. However, we expect that a labeling expert with profound knowledge of package leaflets would be able to generate (with some difficulty) satisfactory text in the large majority of cases. Successful generation thus requires the model to learn specific syntax, terminology, and writing style from the corpus (e.g., via fine-tuning).

\begin{table*}[ht!]
\centering
\begin{tabular}{l l l l l l l}
\hline
\hline
\multirow{2}{4em}{Model} &  
\multicolumn{2}{c}{Word-overlap metrics} & \multicolumn{3}{c}{Semantic equivalence
metrics} \\
& SacreBLEU & ROUGE-L & BERTScore & BLEURT & MoverScore-2l \\
\hline
Content Planner & \textbf{27.78} & 39.32 & 0.214 & -0.072 & 0.591 \\
BART-base    & 8.76 $ \pm $ 0.02  & 42.73 $ \pm $ 0.11 & \textbf{0.370} $ \pm $ 0.001 &  \textbf{0.268} $ \pm $ 0.002 & 0.609 $ \pm $ 0.0004  \\
BART-base + cond  & 8.73 $ \pm $ 0.02 & 42.60 $ \pm $ 0.12 & \textbf{0.369} $ \pm $ 0.001 &  \textbf{0.268} $ \pm $ 0.003 & 0.608 $ \pm $ 0.0004  \\
T5-base         & 18.68 $ \pm $ 0.07 & \textbf{47.22} $ \pm $ 0.17 & 0.363 $ \pm $ 0.001 &  0.255 $ \pm $ 0.008 & \textbf{0.620} $ \pm $ 0.0005 \\
T5-base + cond    & 18.63 $ \pm $ 0.14 & \textbf{47.31} $ \pm $ 0.22 & 0.364 $ \pm $ 0.002 &  0.256 $ \pm $ 0.006 & \textbf{0.621} $ \pm $ 0.0008 \\
\hline
\hline
\end{tabular}
\caption{\label{benchmark-table} Results on the BioLeaflets test set (averaged over all sections). T5 and BART models are fine-tuned with seven different random seeds: average and standard deviation are reported. BLEURT-large-128 is used.}
\end{table*}

\begin{table*}[ht!]
\centering
\begin{tabular}{C{0.15\textwidth} C{0.15\textwidth} C{0.1\textwidth} C{0.15\textwidth} C{0.12\textwidth} C{0.1\textwidth}}
\hline
\hline
Model & & Adequacy & Hallucination presence & Entity inclusion & Fluency \\
\hline
\multirow{2}{*}{Content Planner} &annotator 1   & 4.1 $ \pm $ 3.0  & 6.8 $ \pm $ 3.2 & 4.8 $ \pm $ 3.2 &  5.1 $ \pm $ 3.3   \\
 &annotator 2    & 3.7 $ \pm $ 2.6  & 6.4 $ \pm $ 2.5 & 5.1 $ \pm $ 2.5 &  5.4 $ \pm $ 2.3   \\
\hline
\multirow{2}{*}{BART-base} &annotator 1    & 7.5 $ \pm $ 2.1  & 3.1 $ \pm $ 2.6 & 7.4 $ \pm $ 2.3 &  8.6 $ \pm $ 1.8   \\
 &annotator 2    & \textbf{6.6} $ \pm $ 2.2  & \textbf{3.3} $ \pm $ 2.1 & \textbf{8.1} $ \pm $ 1.8 &  8.0 $ \pm $ 1.3   \\
\hline
\multirow{2}{*}{T5-base} & annotator 1    & \textbf{7.8} $ \pm $ 1.8  & \textbf{3.0} $ \pm $ 2.4 & \textbf{7.6} $ \pm $ 2.1 &  \textbf{9.0} $ \pm $ 1.4   \\
 &annotator 2    & 6.5 $ \pm $ 2.2  & 3.5 $ \pm $ 1.9 & 7.8 $ \pm $ 1.7 &  \textbf{8.2} $ \pm $ 1.2   \\
\hline
\hline
\end{tabular}
\caption{\label{human-evaluation-table} Human evaluation of test samples. Values on a scale from one to ten; average and standard deviation are reported. The higher the better for all quantities, expect for ``Hallucination presence".  Adequacy estimates the overall generation quality, taking into consideration fluency, amount of hallucination, and entities included in the generated text.}
\end{table*}

\section{Experiments} \label{sec:Experiments}

Following \citet{Kale:20}, we represent the structured data (i.e., detected entities) as a flat string (linearization). The entities are kept in their order of appearance (Table\ref{generations-table}b). 
The models are then trained to predict - starting from these entities - the corresponding published leaflet text.

We present baseline results on \emph{BioLeaflets} dataset by employing the following state-of-the-art approaches: 

\begin{itemize}

\item \textbf{Content Planner}: two stages neural architecture (content selection and planning) based on LSTM ~\citep{Puduppully:19}.
Since only relevant entities are provided as input to the model, we solely use the content planning stage (encoder-decoder architecture with an attention mechanism). We train one model for each section, and use the same hyperparameters reported by \citet{Puduppully:19}. 

\item \textbf{T5}: a text-to-text transfer transformer model ~\citep{T5_model:20}. \citet{Kale:20}: showed that T5 outperforms alternatives like BERT ~\citep{Devlin:19} and GPT-2 ~\citep{radford2019language}. 
After hyperparameter search on the development dataset, the following parameters (yielding the best ROUGE-L score ~\citep{lin-2004-rouge}) are selected: constant learning rate of 0.001, batch size of 32, 20 epochs, greedy search as a decoding method. 

\item \textbf{BART}: denoising autoencoder for pretraining sequence-to-sequence models with transformers ~\citep{Lewis:20}. For computational reasons, we use the same hyperparameters as per T5 fine-tuning.

\item \textbf{BART and T5 with conditioning}: we add the prefix “section\_$n$” ($n=1, \dots 6$) to the (linearized) input data. This explicitly gives the model information on the section number and thus enforces a conditioning on the section type for text generation.

\end{itemize}

BART and T5 fine-tuning are performed via HuggingFace ~\citep{wolf-etal-2020-transformers}. 

\section{Evaluation} \label{sec:Evaluations}

Table \ref{generations-table} shows the generated text for one test sample as illustrative example. All generated text is made available\footnote{\url{https://github.com/bayer-science-for-a-better-life/data2text-bioleaflets}}. After a thorough inspection of the samples, we conclude that generated text is generally fluent and coherent. Text produced by T5 and BART is more fluent, factually and grammatically correct than those by Content Planner.
Table \ref{benchmark-table} illustrates the performance of state-of-the-art models quantified by automatic metrics.
\begin{figure*}[t!]
\centering
\includegraphics[width=\textwidth]{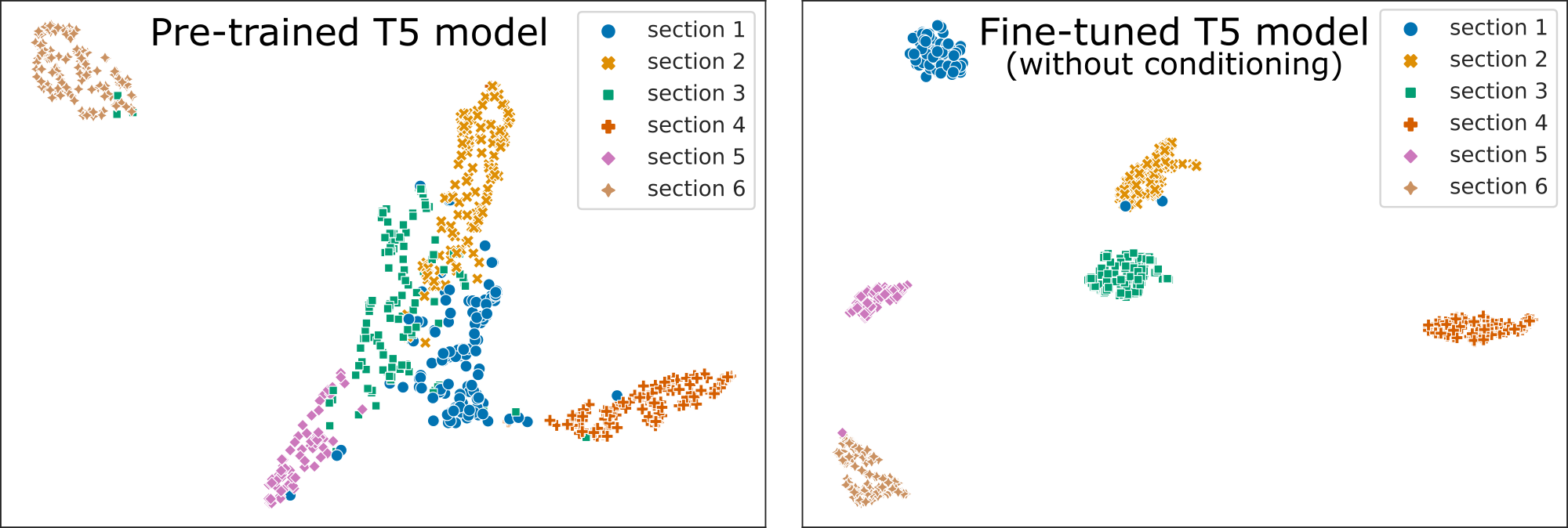}
\caption{Two-dimensional projections of T5 internal representations (average of the last encoder hidden-states) for pre-trained (not fine-tuned) (\textbf{left}) and fine-tuned T5 model on \emph{BioLeaflets} dataset (\textbf{right}). T5 implicitly learns to condition on section type during fine-tuning.}
\label{figure-clustering-section}
\end{figure*}
Word-overlap metrics such as (Sacre)BLUE ~\citep{SacreBLEU:18} and ROUGE \citep{lin-2004-rouge} have been shown to perform poorly in evaluation of natural language generation \citep{novikova-etal-2017}, and thus we report them here only for completeness.
Conversely, contextual embedding based metrics BERTScore ~\citep{bert-score}, BLEURT ~\citep{bleurt-score}, and MoverScore-2 ~\citep{moverscore:19} correlate with human judgment on sentence-level and system-level evaluation. They adequately capture semantic equivalence between generated and target text as well as fluency and overall quality.
T5 and BART outperform Content Planner, as measured by BERTscore, BLEURT, and MoverScore-2. T5 and BART show similar performance.
These results show that transformer-based models and transfer learning strategies achieve state-of-the-art performance on data-to-text tasks, generalizing the findings in  \citet{Kale:20} to multi-sentence and multi-section generation, biomedical text, and low-data setting.

To confirm these findings, human evaluation is performed for Section 1 of the test set by two annotators. Results are shown in Table \ref{human-evaluation-table}.
Similarly to \citet{human-annotation-Manning20}, we design a survey which includes adequacy (estimate of overall quality), presence of hallucinations, entity inclusion, and fluency. 
%
T5 and BART have similar performance, and they produce more adequate text than Content Planner. T5 and BART performance is more stable across samples (lower standard deviation). These conclusions coincide with the ones drawn from Table \ref{benchmark-table}, thus confirming the usefulness of semantic equivalence metrics for automatic evaluation of text generation.

Interestingly, specifying the section type in the input records (i.e., explicit conditioning) did not improve model performances (Table \ref{benchmark-table}). 
To rationalize this result, we analyze T5 internal representations. Specifically, for each test sample, we extract the (average) last encoder hidden-state for both pre-trained (not fine-tuned) and fine-tuned T5 (fine-tuned on \emph{BioLeaflets} but without explicit conditioning). We then project these vectors into two-dimensions using the non-linear dimensionality reduction method UMAP ~\citep{mcinnes2020umap}. The results are depicted in Fig. \ref{figure-clustering-section}.
In Fig. \ref{figure-clustering-section} (right), we can identify six well-separated clusters, which correspond to (the internal representations of samples belonging to) the six document sections in the \emph{BioLeaflets} dataset. Thus, after fine-tuning, T5 maps input data belonging to different sections to different parts of the internal representation space.
The cluster separation is much less pronounced for the pre-trained (not-fine-tuned) T5 model (Fig. \ref{figure-clustering-section}, left).
This shows that during the fine-tuning process, T5 implicitly learns to condition on section type, thus learning to generate different sections, even despite the small dataset. 
Since conditioning is learned automatically, explicitly passing the section type as input does not increase model performance.

\section{Error Analysis and Limitations} \label{sec:limitations}
After thorough qualitative evaluation of numerous generated samples, the following general issues appear: 

\begin{itemize}

\item \textbf{Typos}: Even though models largely utilize the input entities correctly, typos appear in generated text by T5 and BART for out-of-vocabulary words, e.g. Table \ref{generations-table} (c, d). Content Planner does not seem to have this problem.

\item \textbf{Hallucinations} are present for all models. Loss functions like maximum likelihood do not directly minimize hallucinations, thus hindering consistent fact-based text generation.

\item \textbf{Repetitiveness}: Content Planner produce repetitions (e.g. Table  \ref{generations-table} (e)), whereas T5 and BART language models do not. 

\item \textbf{Difficulties in producing coherent long text}: In the \emph{BioLeaflets} dataset, models perform well in generating section 1, which is 962 characters long on average. However, the quality of section 4 "Possible side effects" (3\,453 characters long on average) generation is poor. 

\end{itemize}

Possible improvements to our work are: analysis of the impact of shuffling of entities for the input data generation,  introduction of loss functions that explicitly favor factual correctness, usage of specialized biomedical embeddings, inclusion of more source input data (e.g. part-of-speech, dependency tag), generation of longer text (beyond the 512 tokens generated here).

\section{Conclusion} \label{sec:NAME}

In this study, we introduce a new biomedical dataset (\emph{BioLeaflets}), which could serve as a benchmark for biomedical text generation models. 
We demonstrate the feasibility of generating coherent multi-sentence biomedical text using patient-friendly language, based on input consisting of biomedical entities.
These results show the potential of text generation for real-world biomedical applications. Nevertheless, human evaluation is still a required step to validate the generated samples. Application of the methodology and models used here to different sets of biomedical text (e.g., generation of selected sections of clinical study reports) could be an area for further research.

\bibliographystyle{acl_natbib}
\bibliography{acl2021}

\end{document}